\documentclass{article}
\usepackage{spconf,amsmath,graphicx}
\usepackage{cases}
\usepackage{multirow, tabularx}
\usepackage{hyperref}
\usepackage{xfrac} 
\usepackage{float}
\usepackage{algorithm}
\usepackage{algpseudocode}
\usepackage{enumitem}
\usepackage{pdfpages}
\usepackage{amssymb}
\usepackage{pifont}

\newcolumntype{C}{>{\centering\arraybackslash}X}
\newcommand\blfootnote[1]{%
  \begingroup
  \renewcommand\thefootnote{}\footnote{#1}%
  \addtocounter{footnote}{-1}%
  \endgroup
}
\restylefloat{table}
\hypersetup{
    urlcolor=blue
}

\title{VizECGNet: Visual ECG Image Network for Cardiovascular Diseases Classification with Multi-Modal Training and Knowledge Distillation}
\name{Ju-Hyeon Nam$^{\star}$ \qquad Seo-Hyung Park$^{\star}$ \qquad Su Jung Kim \qquad Sang-Chul Lee}
\address{Department of Electrical and Computer Engineering, Inha University, Incheon, Republic of Korea}
\begin{document}
%
\maketitle
\begin{abstract}
An electrocardiogram (ECG) captures the heart's electrical signal to assess various heart conditions. In practice, ECG data is stored as either digitized signals or printed images. Despite the emergence of numerous deep learning models for digitized signals, many hospitals prefer image storage due to cost considerations. Recognizing the unavailability of raw ECG signals in many clinical settings, we propose VizECGNet, which uses only printed ECG graphics to determine the prognosis of multiple cardiovascular diseases. During training, cross-modal attention modules (CMAM) are used to integrate information from two modalities - image and signal, while self-modality attention modules (SMAM) capture inherent long-range dependencies in ECG data of each modality. Additionally, we utilize knowledge distillation to improve the similarity between two distinct predictions from each modality stream. This innovative multi-modal deep learning architecture enables the utilization of only ECG images during inference. VizECGNet with image input achieves higher performance in precision, recall, and F1-Score compared to signal-based ECG classification models, with improvements of 3.50\%, 8.21\%, and 7.38\%, respectively.
\end{abstract}
\blfootnote{$\star$ denotes the same contributions.}
\blfootnote{This work was supported in part by the National Research Foundation of Korea (NRF) under Grant NRF-2021R1A2C2010893 and in part by Institute of Information and communications Technology Planning \& Evaluation (IITP) grant funded by the Korea government(MSIT) (No.RS-2022-00155915, Artificial Intelligence Convergence Innovation Human Resources Development (Inha University).}
\section{Introduction} 
\label{sec:intro}

Cardiovascular disease ranks as the second leading cause of death, following cancer. Physicians  employ electrocardiograms (ECGs) as a diagnostic tool to monitor the heart's electrical activity. During an ECG, electrodes are placed on the skin, typically on the chest, arms, and legs, to detect and record the heart's electrical impulses. Healthcare professionals analyze these signals to diagnose various heart conditions, including arrhythmias, myocardial infarction (heart attack), and heart failure. Despite its utility, manual analysis of ECG signals is complex and prone to human error, potentially leading to the oversight of subtle yet critical diagnostic patterns.

With advancements in deep learning and signal processing, researchers have strived to automate cardiovascular disease diagnosis. For instance, \cite{xiong2017robust} introduced a binary classification model for arrhythmic fibrillation signals, utilizing a 1D convolutional neural network (CNN) based on single-lead ECG data. Subsequently, RhythmNet \cite{xiong2018ecg} and Stacked CNN-LSTM \cite{tan2018application} addressed the periodic nature of normal signals by incorporating recurrent neural networks (RNN and LSTM) to detect abnormalities based on long-term dependencies. Additionally, \cite{rubin2018densely} proposed a method that extracts multi-scale features to classify diseases from abnormal ECG signals.

In clinical practice, physicians generally prefer utilizing multi-lead ECG signals, which are gathered from multiple electrodes, over single-lead ECG signals for diagnostic purposes. Consequently, numerous researchs have focused on developing diagnostic tools based on multi-lead signals. For example, \cite{chen2019large} employed residual learning and recurrent neural networks to classify seven abnormal signals from 12-lead ECG data. Similarly, \cite{liu2018automatic} classified nine abnormal signals by combining new features obtained by experts from 12-lead ECG signals with features extracted from 1D CNN. Addressing the challenge posed by the concurrent manifestation of multiple cardiovascular diseases in a 12-lead ECG signal, 1D RANet \cite{liu2021multi} tackles a multi-label classification task.

\begin{figure*}[t]
    \centering
    \includegraphics[width=\textwidth]{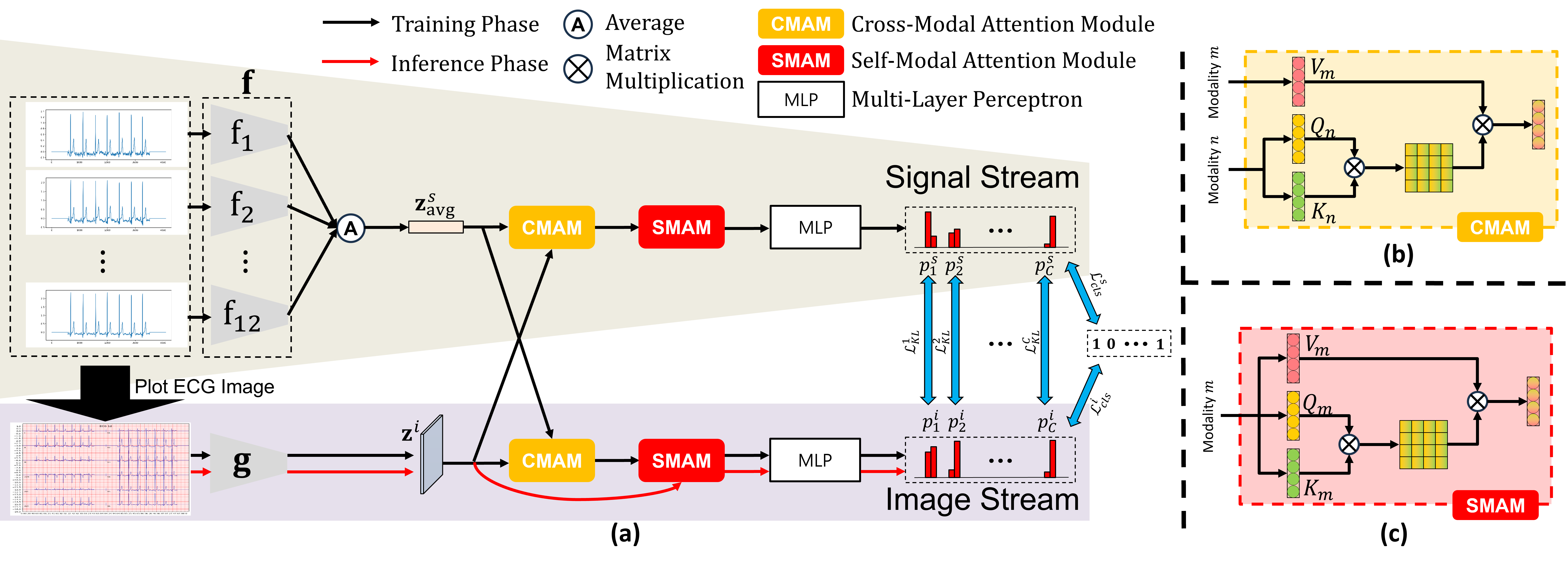}
    \caption{Overall architecture of the proposed VizECGNet, which mainly comprises CMAM and SMAM. (a) Overall block diagram of out network. (b) Overview of CMAM. (c) Overview of SMAM.} 
    \label{fig:VizECGNet}
\end{figure*}

We observed that ECG signal classification models primarily rely on digitized signals for identifying abnormalities. Our primary motivation arises from the unavailability of digitized ECG signals, particularly in smaller clinics, due to factors such as the high maintenance costs of databases and the use of legacy devices. With this motivation, we introduce VizECGNet, a model for classifying cardiovascular diseases using printed ECG graphics (images). We utilized features extracted from both the image and the signal. Recognizing that features from each modality exhibit distinct characteristics, we apply a cross-modal attention module (CMAM) to fuse these features. Additionally, a self-modal attention module (SMAM) refines the extracted features across heterogeneous modalities, emphasizing discriminative features crucial for distinguishing normal and abnormal signals. These modality-specific features are then integrated in fully-connected layers, with knowledge distillation applied between the two predictions to prevent performance degradation when utilizing only images during inference. In comparison to existing models, our approach demonstrates superior classification performance across various metrics such as precision, recall, and F1-Score on large-scale ECG datasets. Our main contribution of this paper is as follows: 

\begin{itemize}
\item We propose a novel ECG multi-label classification model (VizECGNet) based on multi-modal learning and knowledge distillation to employ the complexity characteristics of time-series data within 12-lead ECG signals while using only images during inference. 

\item We also propose two attention modules, CMAM and SMAM, enabling the model to exchange information between modalities and emphasize the discriminative features in each modality stream. 

\item We experimentally achieved state-of-the-art performance in various evaluation metrics (precision, recall, and F1-Score) when comparing signal-/image-/hybrid-based classification models on a large-scale 12-lead ECG dataset. 
\end{itemize}

\section{Method} 
VizECGNet is a composition of 1D and 2D CNN for multi-label classification of 12-lead ECG signals. Our model is a novel structure that combines multi-modal learning and knowledge distillation techniques. For multi-modal learning, we adopt a self-attention mechanism between different and the same modality, called CMAM and SMAM. The features of each modality are forwarded through fully-connected layers for knowledge distillation. Only ECG images are used for predicting cardiovascular disease during the inference phase in 12-lead ECG signals. Fig. \ref{fig:VizECGNet} illustrates the overall structure of VizECGNet.

\subsection{Cross- and Self-Modal Attention Modules} 
The main goal of VizECGNet is to extract the correlations between different modalities and extract discriminative features via cross- and self-modal attention modules. To achieve this goal, we extract features from the 12-lead ECG signals $\mathbf{X}^{s} = \{ \mathbf{x}^{s}_{1}, \dots, \mathbf{x}^{s}_{12} \}$ and image $\mathbf{X}^{i}$ using CNN-based feature extractors $\mathbf{f}$ and $\mathbf{g}$, respectively. For each $l$-th single-lead signal $\mathbf{x}^{s}_{l} = [x^{s}_{l, 1}, \dots, x^{s}_{l, T}]$ with time length $T$, we use twelve different 1D CNN-based feature extractor $\mathbf{f} = \{ f_{1}, \dots, f_{12} \}$ with sharing parameters to efficiently fuse each signal features as follows:
\begin{equation}
\mathbf{z}^{s}_{\text{avg}} = \frac{1}{12} \sum_{l = 1}^{12} f_{l} (\mathbf{x}^{s}_{l}) \in \mathbb{R}^{C^{s} \times T^{s}} .
\end{equation}
Similar to ECG signal $\mathbf{X}^{s}$, we extract the feature maps from 12-lead ECG image $\mathbf{X}^{i}$ using 2D CNN-based feature extractor $\mathbf{g}$ as follow:
\begin{equation} 
\mathbf{z}^{i} = \mathbf{g} (\mathbf{X}^{i})  \in \mathbb{R}^{C^{i} \times H^{i} \times W^{i}} ,
\end{equation}

where $C^{s} = C^{i} = 512$. Subsequently, we apply cross-modal attention module between two extracted features $\mathbf{z}^{s}_{\text{avg}}$ and $\mathbf{z}^{i}$. For clarify, let assume we use self-attention on modality $m$ based on modality $n$. Then, we can write CMAM as follows:
\begin{equation}
\bar{\mathbf{z}}^{m \leftarrow n} = \text{Softmax} \left( \mathbf{Q}_{n} \mathbf{K}^{T}_{n} \right) \mathbf{V}_{m}\\
\end{equation}

where $\mathbf{Q}_{n} = W_{\mathbf{Q}_{n}}\mathbf{z}^{n}, \mathbf{K}_{n} = W_{\mathbf{K}_{n}}\mathbf{z}^{n}, \mathbf{V}_{m} = W_{\mathbf{V}_{m}}\mathbf{z}^{m}$. Then, each refined features $\bar{\mathbf{z}}^{s \leftarrow i}$ and $\bar{\mathbf{z}}^{i \leftarrow s}$ contain different modality information. Now, we apply SMAM to extract discriminative features for two features with mixed information in each modal stream as follows:
\begin{equation}
\begin{cases}
&\hat{\mathbf{z}}^{s} = \textbf{Softmax} \left( \mathbf{Q}_{s} \mathbf{K}^{T}_{s} \right) \mathbf{V}_{s}\\
&\hat{\mathbf{z}}^{i} = \textbf{Softmax} \left( \mathbf{Q}_{i} \mathbf{K}^{T}_{i} \right) \mathbf{V}_{i}
\end{cases}
\end{equation}

where $\mathbf{X}_{m} = W_{\mathbf{X}_{m}}\bar{\mathbf{z}}^{m \leftarrow n}$ for $m \in \{ s, i \}$ and  $\textbf{X} \in \{ \textbf{Q}, \textbf{K}, \textbf{V} \}$. Each final refined discriminative features $\hat{\mathbf{z}}^{s}$ and $\hat{\mathbf{z}}^{i}$ are forwarded into the fully-connected layers for each modality stream to classify abnormal ECG signal types.

\subsection{Knowledge Distillation} 
Knowledge distillation is a technique commonly used in machine learning to transfer knowledge from a complex model (teacher model) to a simple model (student model). To utilize only printed ECG signal to classify abnormal signals during inference and acquire knowledge about 12-lead ECG signals, we adopt knowledge distillation from signal stream into image stream. First, predicting the probability distributions $p^{s}$ and $p^{i}$ for each class is performed using a classifier for each modality stream as follows:
\begin{equation}
    p^{m} = \textbf{MLP}_{m} \left( \textbf{GAP} \left( \hat{\mathbf{z}}^{m} \right) \right)
\end{equation}

where $\textbf{MLP}_{m} ( \cdot )$ is a multi-layer perceptron (MLP) for each signal and image modality stream. For each prediction, the classification loss function $\mathcal{L}_{cls}$ for the same label $t$ is computed as follows:
\begin{equation}
\mathcal{L}_{cls} = \sum_{c = 1}^{C} \left( \mathcal{L}_{BCE} \left( t_{c}, p^{s}_{c}  \right) + \mathcal{L}_{BCE} \left( t_{c}, p^{i}_{c}  \right) \right)
\end{equation}

where $C$ is a number of classes and $\mathcal{L}_{BCE}$ is the binary cross-entropy loss function. Since the dataset used in this paper can have multiple diseases for a single ECG signal, a binary classification is performed for each class. Finally, we calculate knowledge distillation loss $\mathcal{L}_{KD}$ between two modality streams to reduce the difference in probability distribution $p^{s}$ and $p^{i}$ as follows:
\begin{equation}
\begin{split}
\mathcal{L}_{KD} (p^{s}, p^{i}) = \sum_{c = 1}^{C} \mathcal{L}_{KL} (p^{s}_{c} || p^{i}_{c}) \\ = \sum_{c = 1}^{C} \sum_{x \in \mathcal{X}} p^{s}_{c}(x) \text{log} \left( \frac{p^{s}_{c} (x)}{p^{i}_{c} (x)} \right) 
\end{split}
\end{equation}

where $\mathcal{L}_{KL}$ is a Kullback-Leibler Divergence to calculate the difference between two probability distributions $p^{s}_{c}$ and $p^{i}_{c}$ for each class $c$. The final loss function $\mathcal{L}_{total} = \lambda_{1} \mathcal{L}_{cls} + \lambda_{2} \mathcal{L}_{KD}$ is used for updating parameters of each modality stream. To reduce the sensitivity of hyperparameters, we fix $\lambda_{1} = \lambda_{2} = 1$.

\begin{table*}[t]
    \centering
    \footnotesize	
    \begin{tabular}{c|c|c|c|c|ccc}
    \hline
    Modality & Method & Parameters (M) & Speed (ms) & Inference Data Type & Precision & Recall & F1-Score \\
    \hline
    \multicolumn{1}{c|}{\multirow{5}{*}{Signal}} & InceptionTime \cite{ismail2020inceptiontime} & 0.49M & 15.9ms & \multicolumn{1}{c|}{\multirow{5}{*}{Signal}} & 46.13 & 39.98 & 42.84 \\
    \multicolumn{1}{c|}{} & XResNet1D-101 \cite{he2019bag} & 13.94M & 23.7ms & & 49.26 & 42.96 & 45.89 \\
    \multicolumn{1}{c|}{} & Transformer \cite{vaswani2017attention} & 24.17M & 32.1ms & & 51.34 & 41.73 & 46.04 \\
    \multicolumn{1}{c|}{} & ACNet \cite{ding2019acnet} & 262.00M & 24.5ms & & 53.83 & 45.12 & 49.09 \\
    \multicolumn{1}{c|}{} & 1D RANet \cite{liu2021multi} & 3.93M & 23.5ms & & 57.73 & 48.51 & 51.51 \\
    \hline
    \multicolumn{1}{c|}{\multirow{2}{*}{Image}} & ResNet18 \cite{he2016deep} & 11.64M & 21.4ms & \multicolumn{1}{c|}{\multirow{2}{*}{Image}} & 52.62 & 47.14 & 49.73 \\
    \multicolumn{1}{c|}{} & MobileNetV3 \cite{howard2019searching} & 18.20M & 12.8ms & & 39.56 & 31.83 & 35.28 \\
    \hline
    \multicolumn{1}{c|}{\multirow{4}{*}{Signal \& Image}} & MHM \cite{qiao2020mhm} & 15.44M & 28.1ms & \multicolumn{1}{c|}{\multirow{2}{*}{Signal}} & 44.16 & 39.42 & 41.68 \\
    \multicolumn{1}{c|}{} & \textbf{VizECGNet (Ours)} & 11.17M & 37.1ms & & \textcolor{red}{\textbf{\underline{63.20}}} & \textcolor{red}{\textbf{\underline{59.11}}} & \textcolor{red}{\textbf{\underline{61.09}}} \\ \cline{2-8}
     \multicolumn{1}{c|}{} & MHM \cite{qiao2020mhm} & 15.44M & 30.2ms & \multicolumn{1}{c|}{\multirow{2}{*}{Image}} & 37.18 & 32.71 & 34.83 \\
    \multicolumn{1}{c|}{} & \textbf{VizECGNet (Ours)} & 11.17M & 39.3ms & & \textcolor{blue}{\textbf{\textit{61.23}}} & \textcolor{blue}{\textbf{\textit{56.72}}} & \textcolor{blue}{\textbf{\textit{58.89}}} \\
    \hline
    \end{tabular}
    \caption{Experiment results on the Large-Scale ECG datasets. We also provide number of trainable parameters (M) and inference speed (sec) for each methods. \textcolor{red}{\textbf{\underline{Red}}} and \textcolor{blue}{\textbf{\textit{Blue}}} are the first and second performance results, respectively.}
    \label{tab:summary} 
\end{table*}

\section{Experimental Results} 
\subsection{Experimental Settings and Implementation Details} 
We implemented VizECGNet in Pytorch 1.11 and Python 3.8. The large-scale 12-lead ECG signal dataset \cite{ribeiro2021code} used in this paper is multi-label (1dAVb, RBBB, LBBB, SB, AF, ST). The ECG signal of each lead is composed of 4096 time lengths. In this paper, to eliminate the trend of each signal, the average voltage for the entire time is set to zero mean, and we apply a detrending process. Finally, we convert the 12-lead ECG signals using the ecg\_plot Python library for multi-modal learning on the images. We compared our VizECGNet with five signal-based models (InceptionTime \cite{ismail2020inceptiontime}, XResNet1D-101 \cite{he2019bag}, Transformer \cite{vaswani2017attention}, ACNet \cite{ding2019acnet}, and 1D RANet \cite{liu2021multi}), two image-based models (ResNet18 \cite{he2016deep} and MobileNetV3 \cite{howard2019searching}), and multi-modal models (MHM \cite{qiao2020mhm}).  Since the use of default training settings generally performs poorly in ECG dataSet, for fair comparison, we optimized the parameters to work best on ECG dataset. We train all models in an end-to-end manner using the Adam optimizer. The initial learning rate starts from $10^{-3}$ and is decreased to $10^{-6}$ using the cosine annealing learning rate scheduler \cite{loshchilov2016sgdr}, and the training settings were set to a batch size of 16 and epochs of 300 till the loss functions of all models converged. For evaluation, we used three metrics (Precision, Recall, and macro-averaged F1-Score) to measure the performance of each model. To efficiently extract features from signals and image, we utilize ResNet18 as feature extractor. 
 
\subsection{Results Analysis} 
As shown in Table \ref{tab:summary}, VizECGNet outperforms single- and multi-modality models on all performance evaluation metrics in both inference data types. When predicting using ECG signals, VizECGNet achieve 9.58\%, and 19.68\% higher F1-Score compared with 1D RANet, and MHM, respectively. Furthermore, when using printed ECG images, VizECGNet achieve 17.21\%, and 24.06\% higher F1-Score compared with ResNet18, and MHM, respectively. Note that MobileNetV3 and ResNet18 used only simple ECG images for training and evaluation. This training strategy makes two image-based models unable to understand the abnormal signal characteristics of each lead. However, although 1D RANet receives 12-lead ECG signal data with complex data structures, its performance is low. These reasons indicate that the interpretability of 1D RANet for complex data is still poor. On the other hand, VizECGNet achieves high classification performance because it exchanges information based on multi-modal learning and distills the knowledge of abnormal signals generated by subtle differences in 12-lead ECG signals into an image modality stream during learning. 

\begin{figure}[t]
    \centering
    \includegraphics[width=0.45\textwidth]{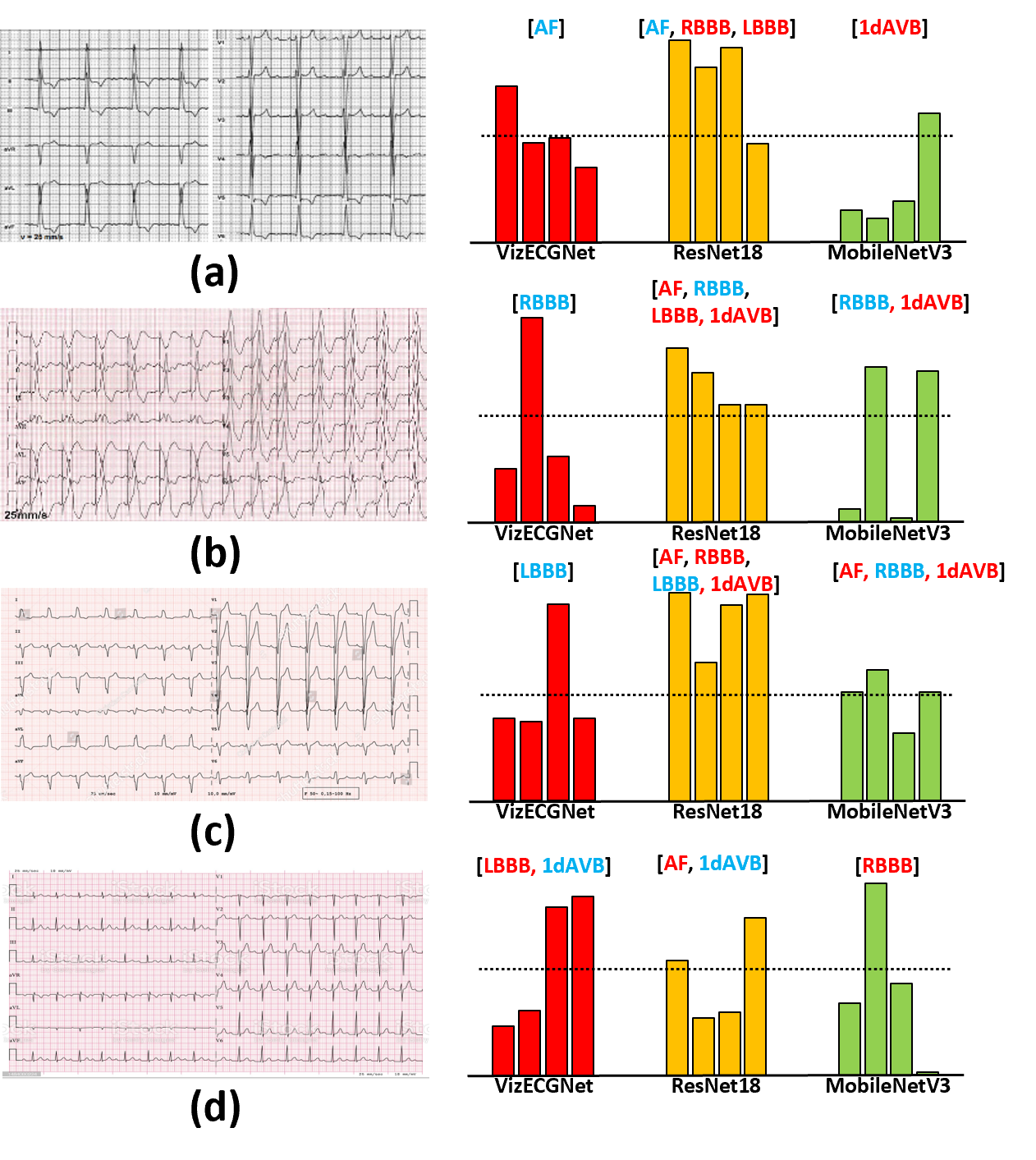} 
    \caption{The example of real ECG print image and prediction results of VizECGNet and Image-based models (ResNet18 and MobileNetV3). (a) \href{http://www.hvt-journal.com/articles/art6}{AF}. (b) \href{https://www.researchgate.net/figure/9-A-12-lead-ECG-showing-RBBB-with-right-axis-deviation-and-positive-precordial_fig4_367943268}{RBBB}. (c) \href{https://www.shutterstock.com/search/left-bundle-branch}{LBBB}. (d) \href{https://www.istockphoto.com/kr/벡터/ecg-1도-방실-차단-1도-방실-차단-12리드-ecg-공통-사례-6초-리드-gm1484000224-510460572}{1dAVB}. (e) Prediction probability for each cardiovascular diseases. \textcolor{red}{\textbf{Red}}, \textcolor{yellow}{\textbf{Yellow}}, and \textcolor{green}{\textbf{Green}} bars denotes VizECGNet, ResNet18, and MobileNetV3, respectively.} 
    \label{fig:practical_results}
\end{figure}

We also examined extrapolation to actual ECG printed images to confirm the utility of the model (Fig. \ref{fig:practical_results}). In addition, the inference results of two image-based models (ResNet18 and MobileNetV3) are added. Fig. \ref{fig:practical_results}.(e) shows the disease probability for each model for each disease (AF, RBBB, LBBB, 1dAVB). Our model was positive for all diseases, but the other two models were negative for all but well-characterized diseases such as 1dAVB. These results suggest that VizECGNet is more practical than other models because it can be applied to real printed images even though it was trained on synthetic ECG images created from 12-lead ECG signals. 

\subsection{Ablation Study} 
In this section, we analyze the effectiveness of two attention modules (CMAM and SMAM). In Table \ref{tab:attention_modules}, the results of the ablation study on attention modules are listed with four configurations. Basically, attention is used to extract discriminative features from messy features. In this paper, important information can be additionally extracted by focusing on each modality or between modalities. It can be seen from the actual experimental results that when the mode is paid attention to, all performance evaluation indicators have achieved high performance. 

\begin{table}[t]
    \centering
    \footnotesize
    \begin{tabular}{c|ccc}
    \hline
    Settings & Precision & Recall & F1-Score \\
    \hline
    \textbf{VizECGNet (+SMA+CMA)} & \textcolor{red}{\textbf{\underline{61.23}}} & \textcolor{red}{\textbf{\underline{56.72}}} & \textcolor{red}{\textbf{\underline{58.89}}} \\
    \hline
    w/o SMA & 59.53 & 49.06 & 53.83 \\
    w/o CMA & \textcolor{blue}{\textbf{\textit{60.12}}} & \textcolor{blue}{\textbf{\textit{53.72}}} & \textcolor{blue}{\textbf{\textit{56.70}}} \\
    w/o SMA \& CMA & 57.85 & 47.90 & 52.37 \\
    \hline
    \end{tabular}
    \caption{Ablation study of VizECGNet on the Large-Scale ECG dataset for attention modules. \textcolor{red}{\textbf{\underline{Red}}} and \textcolor{blue}{\textbf{\textit{Blue}}} are the first and second performance results, respectively.}
    \label{tab:attention_modules}
\end{table}

\section{Conclusion} 
We propose VizECGNet, which applies multi-modal learning-based knowledge distillation techniques to classify abnormal electrical signals in ECG signals. Experimental results on a large-scale ECG dataset demonstrate that VizECGNet performs better than traditional heart disease classification models. In the multi-modal case, cross- and self-modality attention modules (CMAM and SMAM) enable us to focus on discriminative features between different modalities and apply knowledge distillation techniques to prevent the performance drop when only images are used during inference. These results prove that the model can be fully exploited in developing countries, which only have access to ECG printers without undergoing the refinement process. To further verify the generalization ability of VizECGNet, we plan to train and evaluate it on various 12-lead datasets, make it into an application, and test it in a real clinical setting.

\bibliographystyle{IEEEbib}
\bibliography{main}

\end{document}